%% file: acl_latex.tex
\documentclass[11pt]{article}

\usepackage[preprint]{acl}

\usepackage{times}
\usepackage{latexsym}

\usepackage[T1]{fontenc}

\usepackage[utf8]{inputenc}

\usepackage{microtype}

\usepackage{inconsolata}

\usepackage{graphicx}

\usepackage{booktabs}
\usepackage{multirow}
\usepackage{geometry}
\usepackage{caption}
\usepackage{xspace}
\usepackage[table]{xcolor}
\usepackage{amsmath}

\usepackage{listings}
\lstdefinestyle{prompt}{
  basicstyle=\footnotesize\ttfamily,
  breaklines=true,
  breakatwhitespace=false,
  breakautoindent=true,
  columns=fullflexible,
  keepspaces=true,
  showstringspaces=false,
  frame=single,
  framesep=5pt,
  xleftmargin=5pt, xrightmargin=5pt,
  rulecolor=\color[gray]{0.5},
  backgroundcolor=\color[gray]{0.97},
  literate=
    {á}{{\'a}}1 {é}{{\'e}}1 {í}{{\'i}}1 {ó}{{\'o}}1 {ú}{{\'u}}1
    {à}{{\`a}}1 {è}{{\`e}}1 {ì}{{\`i}}1 {ò}{{\`o}}1 {ù}{{\`u}}1
    {â}{{\^a}}1 {ê}{{\^e}}1 {î}{{\^i}}1 {ô}{{\^o}}1 {û}{{\^u}}1
    {ä}{{\"a}}1 {ë}{{\"e}}1 {ï}{{\"i}}1 {ö}{{\"o}}1 {ü}{{\"u}}1
    {ñ}{{\~n}}1 {ç}{{\c c}}1
    {—}{{\textemdash}}1 {–}{{\textendash}}1
}

\title{Knowledge Pull Requests for Continual Document Authoring}

\author{
Alexander Martin 
\quad Benjamin Van Durme
\\
  Johns Hopkins University \\
  \texttt{\small{\{amart233, vandurme\}@jhu.edu}}
}

\begin{document}
\maketitle
\begin{abstract}
\input{sections/00-abstract}
\end{abstract}

\section{Introduction}
\input{sections/10-intro}

\section{Related Work}
\input{sections/20-related}

\section{Knowledge Pull Requests}
\input{sections/30-kpr}

\input{figures/claim_proposal}
\section{Method for KPRs}

\input{sections/40-method}

\section{Revising Cross-lingual Knowledge Across Wikipedia with KPRs}

\input{sections/50-wiki-exps}

\section{Updating Query-Driven Report Generation with KPRs}
\input{sections/60-ragtime-exps}

\section{Findings Across Settings}
\input{sections/70-findings}

\section{Conclusion}

\input{sections/80-conclusion}

\bibliography{anthology,custom}

\appendix

\section{Wikipedia Experiments Implementation Details}
\label{append:wiki-details}

\input{appendix/wiki-details}

\section{RAGTIME Experiments Implementation Details}
\label{append:ragtime-details}

\input{appendix/ragtime-details}

\section{Method Details}
\label{append:method}

\input{appendix/method-details}

\section{Metric Details}
\label{append:metric}
\input{appendix/metric-details}

\section{Acknowledgement of AI}
We use AI for coding and to edit and condense writing.

\input{appendix/prompts}

\end{document}

%% file: sections/00-abstract.tex
We introduce Knowledge Pull Requests~(KPRs), a framework for continual document authoring that makes each change interpretable. Documents require ongoing revision as new knowledge surfaces from other sources, languages, or times, but existing approaches either edit with no account of what knowledge changed or regenerate from scratch. A KPR integrates new knowledge into a document by extracting claims, filtering and routing them to sections, and flagging conflicts with existing content, producing a ChangeLog that separates what knowledge changes (claim proposal) from how the text changes (document diff). We evaluate KPRs on revising Wikipedia across languages and updating query-driven reports on RAGTIME. KPRs integrate more information and better preserve existing content than rewriting from sources or regenerating from scratch, while adding the most information per token generated. A KPR-revised article also grounds question answering better than a frontier model with search, which does not surface knowledge documented only in other languages.\footnote{\url{https://github.com/alexmartin1722/kpr}}

%% file: sections/10-intro.tex
Many important documents maintained by humans and systems are never truly finished. Wikipedia articles, technical documentation, intelligence reports, and more require ongoing maintenance as knowledge surfaces from new sources, other languages, or later time periods. To incorporate these changes, one may prompt a language model with an existing document alongside new sources \cite{iv-etal-2022-fruit, reddy2025winellwikipedianeverendingupdating}, but this gives no account of what facts are added or overwritten. Query-driven systems do not attempt updating, instead generating documents from scratch for each query \cite{shao-etal-2024-assisting, openai2025deepresearch, google2025geminidr}. If the query is asked again or new sources surface, the entire generation pipeline reruns.

Even Wikipedia, the canonical continually authored knowledge base, tracks edits only as raw text diffs, recording which words changed and not what knowledge did. With millions of edits per month and a finite pool of volunteers, reviewers struggle to assess whether an edit introduces new facts, corrects outdated ones, or conflicts with existing content \cite{franzmeyer-etal-2024-hellofresh}.

\input{figures/teaser}
Reducing this burden requires moving from text-level diffs to knowledge-level proposals --- an interpretable account of what claims are being added and how they relate to existing content. Software engineering solved an analogous problem with the \emph{Pull Request}, a  proposal of changes to a code base that can be inspected, discussed, and merged. We introduce \emph{Knowledge Pull Requests} (KPRs), the equivalent for documents (\autoref{fig:teaser}). A KPR extracts claims from a set of \emph{sources}, filters and routes them into a \emph{main document}, and flags conflicts, producing a \emph{ChangeLog}: a reviewable artifact separating what knowledge changes (claim proposal) from how the text changes (document diff).

We study continual document authoring in two settings: revision, adding information about an unchanged world, and update, synchronizing a document with a changed world \cite{katsuno1991difference}. For revision, we integrate knowledge into an English Wikipedia article from its counterparts in other languages, where coverage is often uneven, so facts documented in one language may be absent from the English article and hard to reach through English search. For update, query-driven reports must be kept current as new sources extend, supersede, or contradict them. We study this with three constructed variants of RAGTIME \citep{lawrie2026overviewtrec2025ragtime}: temporal, conflict, and balanced.

We compare KPRs against three baselines. The first rewrites the document conditioned on the new sources, but never makes explicit what an edit adds or overwrites \cite{reddy2025winellwikipedianeverendingupdating}. The second conditions on claims extracted from those sources, but without a claim proposal to filter them. The third regenerates the document from scratch over all sources, discarding the existing one. We evaluate these methods intrinsically, by the faithfulness and completeness of the rewrite, extrinsically, by its usefulness as a grounding source for downstream question answering, and by the cost of adding and reviewing its changes.

Across both settings, KPRs outperform these baselines, integrating more of the new information and better preserving the existing document. When revising Wikipedia, the revised article is a stronger grounding source than the original, any baseline, or a frontier model with web search, which does not surface this knowledge. When updating RAGTIME, KPRs flag conflicts between sources rather than silently resolving them, holding precision where conditioning on raw text is misled. Across tasks, we find that conditioning on claims rather than raw source text yields more complete documents. Additionally, filtering those claims before the rewrite raises both precision and recall, while concentrating the changes into contiguous edits a reviewer can approve.

We contribute: (1) \textbf{Knowledge Pull Requests}, which add knowledge to a document as a reviewable \emph{ChangeLog}, and (2) evidence that KPRs outperform existing document-updating approaches.

%% file: figures/teaser.tex
\begin{figure}
    \centering
    \includegraphics[width=\linewidth]{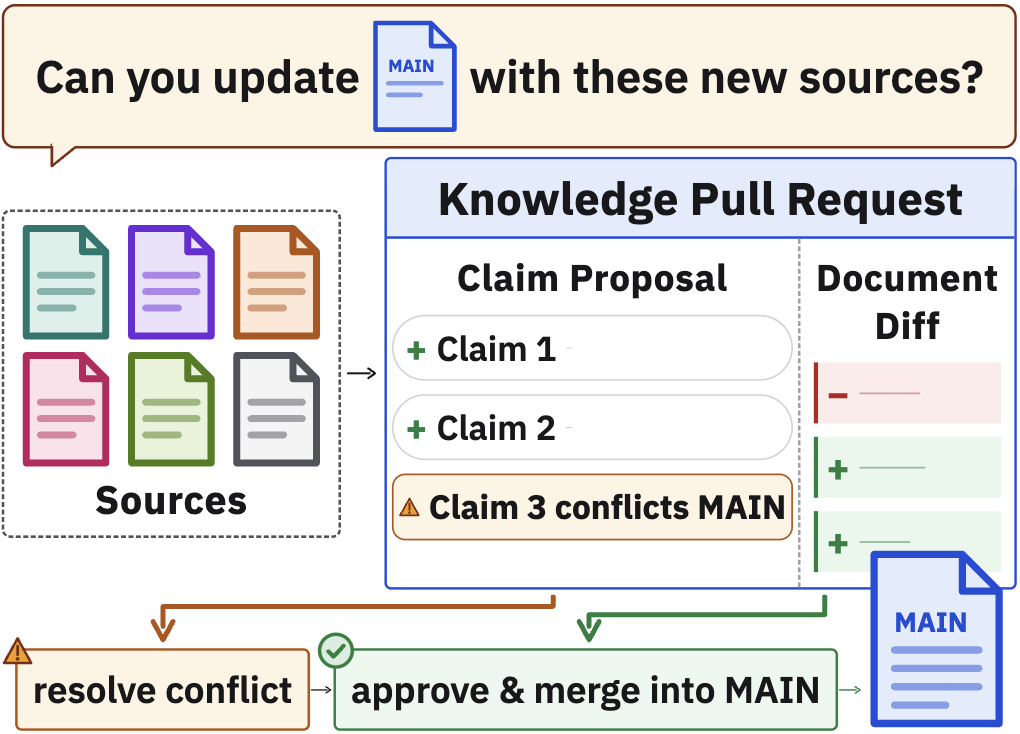}
    \caption{Knowledge Pull Requests update a main document with information from sources as a reviewable ChangeLog: a Claim Proposal (the absent or conflicting source claims) and a Document Diff (the text changes).}
    \label{fig:teaser}
    \vspace{-1.5em}
\end{figure}

%% file: sections/20-related.tex
\paragraph{Belief Revision and Epistemology of Documents.}
Formal epistemology studies how to incorporate new information into an existing body of belief \citep{gardenfors1988knowledge,ferme2011agm}. The AGM framework  \citep{alchourron1985logic} operates over belief sets closed under logical consequence, but requiring the belief state to contain every sentence its members entail is poorly suited to a document. This is refined by \emph{belief bases} \citep{hansson1992defense,hansson1999textbook}, which model finite sets of explicitly held sentences and move closer to how belief is expressed in language. Belief revision further distinguishes \emph{revision}, learning more about a static world, from \emph{update}, where the world itself changes \citep{katsuno1991difference}. These frameworks, however, model belief change over sentences and their logical relations alone, without reference to the justifications behind them, operating at too high a level of abstraction \citep{pollock1987defeasible,pollock2000belief}. Our KPRs take this a step further, operating over natural-language claims rather than formal sentences and producing a ChangeLog that separates the knowledge-level claim proposal from the text-level diff. Our experiments cover both settings: revising English Wikipedia from other-language editions (revision) and updating reports as sources evolve~(update).

\paragraph{Knowledge Cutoffs and Editing.}
A language model's knowledge is fixed at a training cutoff and grows stale as the world changes \citep{jang-etal-2022-temporalwiki, cheng2024dated, vu-etal-2024-freshllms}. A common approach updates the model's parameters: either locating and overwriting factual associations in the weights \citep{meng2022locatingeditingfactualassociations,meng2023massediting} or isolating knowledge in modular adapters that can be added, removed, or swapped \citep{pfeiffer-etal-2021-adapterfusion,fleshman2025adapterswapcontinuoustrainingllms,fleshman2025loraaugmentedgenerationlagknowledgeintensive}. A complementary line instead updates the source text models rely on. Here Wikipedia is a natural target: it is a standard ingredient of LLM pretraining corpora \citep{soldaini-etal-2024-dolma,groeneveld-etal-2024-olmo} and among the most frequently cited domains grounding the answers of LLMs and AI search \citep{semrush2025mostcited,thestacc2026wikipedia,wmf2023value,wmf2025trends}. \citet{iv-etal-2022-fruit} and \citet{reddy2025winellwikipedianeverendingupdating} rewrite Wikipedia articles to reflect new sources, and \citet{spangher-etal-2022-newsedits} model the revision histories of news articles. However, editing weights and text alike treat updating as end-to-end
rewriting, yielding a revised artifact without surfacing which claims changed or how
they conflict with existing content, and offering a reviewer no interpretable
record of the update.

\paragraph{Claims and Factuality.}
Claims---atomic, independently verifiable propositions---have become a standard unit for reasoning about factual content, for two reasons. First, claims are more interpretable than sentences. Building on the Pyramid method \citep{nenkova-passonneau-2004-evaluating}, which scores content by atomic facts (Summary Content Units) rather than whole sentences, FActScore \citep{min-etal-2023-factscore} and VeriScore \citep{song-etal-2024-veriscore} decompose model outputs into subclaims and verify each against evidence, showing that atomic claims support more reliable and interpretable factuality judgments than sentence- or passage-level assessment. Second, claims are more universal and independent of the form of their context than task-specific alternatives \citep[e.g., nuggets;][]{voorhees2003overview}, transferring even to other modalities \citep{jing-etal-2024-faithscore,martin2026seeingmirageevaluatingmultimodal}. These two properties motivate operating over claims rather than whole documents. Claims expose exactly which fact is being asserted, distinguishing our approach from methods that update articles conditioned on raw text \citep{reddy2025winellwikipedianeverendingupdating}. Reducing sources to claims first makes explicit what knowledge each change contributes.

%% file: sections/30-kpr.tex
A Knowledge Pull Request (KPR) is a structured process for integrating new knowledge from a set of source documents into a main document. We define the core components below.

\paragraph{Main Document and Sources.} A KPR operates over two inputs: a \textit{main document} and a set of \textit{sources}. The main document (hereafter, \textit{main}) is a previously written document on a given topic. The sources are documents containing potentially new knowledge relevant to the main's topic. The goal of a KPR is to integrate information from the sources into the main, working with its existing content rather than overwriting it.

\paragraph{Claims.} KPRs operate at the level of \textit{claims}: atomic, decontextualized factual statements \citep[see][]{gunjal-durrett-2024-molecular} extracted from the sources and the main. Operating at the claim level, rather than the passage level, enables precise tracking of what knowledge is being proposed, where in the main it should be placed, and whether it conflicts with existing content.

\paragraph{ChangeLog.} The key artifact produced by a KPR is a \textit{ChangeLog}: a structured, reviewable record of the proposed changes to the main. A ChangeLog consists of two components:

\textit{Claim Proposal.} A mapping of new claims extracted from the sources to the sections of the main document\footnote{We use sections throughout, but a KPR can operate over any granularity of the main (sentences, paragraphs, or pages).} where they are proposed to be added, including new ones where needed. A candidate claim is mapped if it is not filtered by coverage (main already contains it) or against the document's authoring criterion (query relevance or authoring guidelines\footnote{\url{https://en.wikipedia.org/wiki/Wikipedia:What_Wikipedia_is_not\#Encyclopedic_content}}). The claim proposal is also where the KPR flags \textit{knowledge conflicts} (merge conflicts) \citep[inter-context conflicts;][]{xu-etal-2024-knowledge-conflicts}: cases where a proposed claim contradicts an existing claim in main. Rather than silently resolving these conflicts, the KPR surfaces them for review~\citep{thorne-etal-2018-fever}. A human reviewer can adjudicate flagged conflicts, approve or reject claims, and reverse filtering decisions made in the proposal.

\input{figures/claim_decomp}

\textit{Document diff.}
A record of the proposed textual changes to the main, showing how the document would read after the proposed claims are integrated. Together, the claim proposal and diff let a reviewer inspect both \textit{what knowledge is being added} (the claim proposal) and \textit{how the document text changes as a result} (the diff).

The ChangeLog is what distinguishes a KPR from a simple rewrite. By separating the knowledge-level proposal from the text-level changes, it enables collaborative continual authoring. A human reviewer can assess proposed claims on their merits, resolve conflicts, and approve or reject changes before they are merged into main, analogous to a code review in software engineering.

%% file: figures/claim_decomp.tex
\begin{figure}
    \centering
    \includegraphics[width=\linewidth]{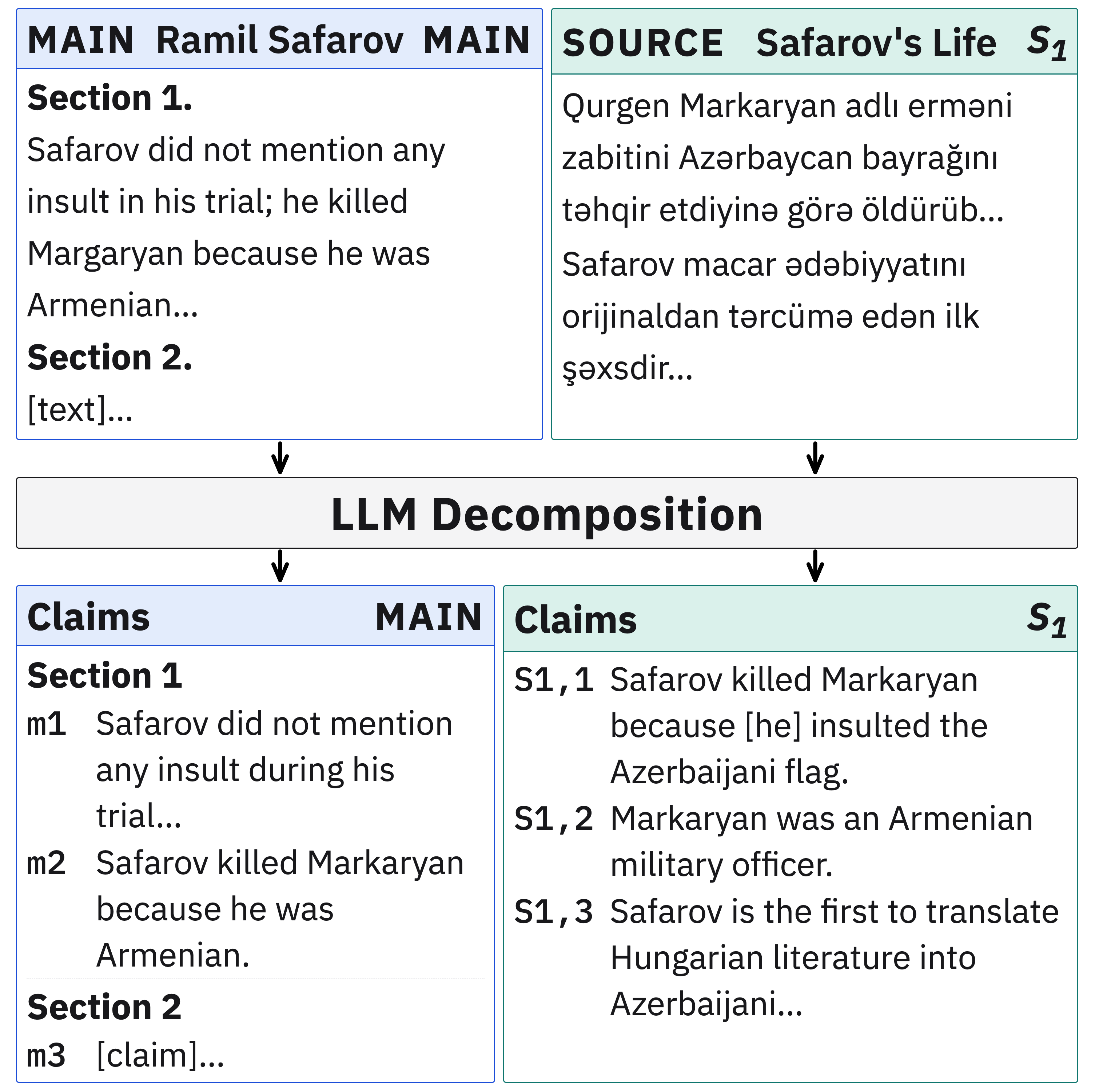}
    \caption{The main and sources are decomposed into atomic, decontextualized claims. Non-English sources are decomposed directly into English, and the main's claims are cached offline.}
    \label{fig:claim_decomp}
    \vspace{-1em}
\end{figure}

%% file: figures/claim_proposal.tex
\begin{figure}
    \centering
    \includegraphics[width=\linewidth]{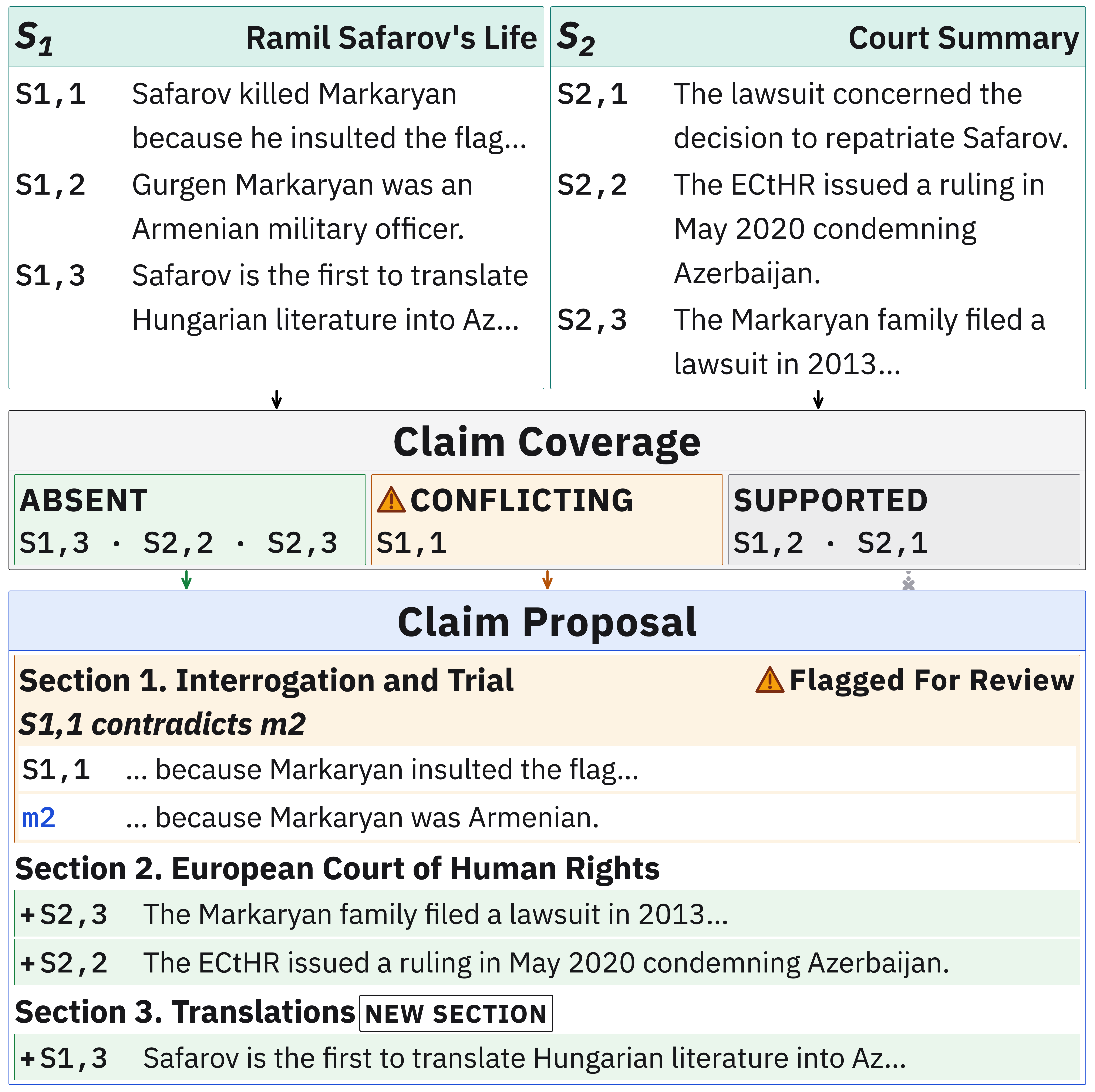}
    \caption{Source claims are classified against the main's claims as absent, conflicting, or supported. Absent claims are filtered for relevance and routed to a section, conflicting claims are flagged for review, and supported and irrelevant claims are dropped.}
    \label{fig:claim_proposal}
    \vspace{-1em}
\end{figure}

%% file: sections/40-method.tex
We introduce a three-stage baseline for producing KPRs. The first stage decomposes the sources into claims. The second produces a claim proposal by filtering claims already covered by or irrelevant to the main, flagging knowledge conflicts, and routing the remainder to sections. Finally, the third rewrites the new and affected sections to produce the diff.

\paragraph{Claim Decomposition (\autoref{fig:claim_decomp}).} 
Using an LLM, we decompose the sources into sets of atomic, decontextualized claims, giving a set of candidate claims to be potentially added to the main. The main is decomposed with the same method, but its claims are decomposed offline and cached as an index rather than online as with the sources. For non-English sources, we decompose directly into English rather than translating first, which yields more faithful claims (\autoref{append:method}).

\paragraph{Claim Proposal (\autoref{fig:claim_proposal}).} 
For each candidate claim, the KPR makes four decisions: (1)~\textit{coverage}, whether the claim is already covered by the main's claims; (2)~\textit{conflict}, whether it contradicts an existing claim in the main; (3)~\textit{relevance}, whether the claim meets the document's authoring criterion; and (4)~\textit{routing}, which section a new, non-conflicting, relevant claim belongs in. Coverage and conflict are resolved in a single classification pass labeling source claims as \textit{covered}, \textit{conflicting}, or \textit{absent}. Relevance is then applied to the absent claims, against the information request in the RAGTIME setting, and left unfiltered in the Wikipedia setting, where candidates already come from articles authored with the same guidelines. Routing maps absent, relevant claims to an existing section or proposes a new one. Covered and irrelevant claims are dropped and conflicting claims are flagged for review.

\paragraph{Document diff (\autoref{fig:doc_diff}).} 
We then apply the claim proposal to the main, one section at a time. Each section (new or existing) that receives one or more claims is rewritten to integrate them, while sections with no proposed claims are left unchanged. Diffing the resulting document against the original main gives the document diff.

\input{figures/doc_diff}

\subsection{Baselines}
We compare KPRs against three methods that integrate the sources without a claim proposal.

\paragraph{ConText.} ConText (Concatenate Text) is our adaptation of WiNELL \cite{reddy2025winellwikipedianeverendingupdating}. It conditions each section's rewrite on raw source text and performs no claim decomposition. In place of WiNELL's retrieval step, an LLM classifies whether a source contains information relevant to that section, also to mirror our claim routing. 

\paragraph{ConClaim.} ConClaim (Concatenate Claims) differs from ConText only in conditioning on source claims rather than source text. Exactly as in KPR, sources are decomposed into claims and those claims are routed to sections. The only difference from KPR is that ConClaim applies no coverage, conflict, or relevance filtering, conditioning each section's rewrite on all claims routed to it. ConClaim is therefore equivalent to a KPR without claim review.

\paragraph{Scratch.} Scratch regenerates the document from all sources in one pass, discarding the standing document entirely. This follows how a query-driven or deep-research system answers an updated query. We use it in the RAGTIME setting only, where regeneration is the standard alternative to updating when new sources surface.

All baselines produce their document diff the same way as KPR, by diffing the rewritten document against the original main.

\subsection{Implementation Details}
All uses of an LLM---classification, routing, and rewriting---use Qwen3.5-27B \citep{qwen35blog}. Because ConText, ConClaim, and KPRs operate over document sections, the two evaluation settings differ in how sections are obtained. For Wikipedia, articles are already organized into sections, so all methods operate on the native section structure. For RAGTIME, system-generated reports do not generally have a section structure, but for our experiments we impose one on the round-1 reports so that every method can operate section by section. A KPR requires only a span the document can be rewritten in, not a pre-existing header, so imposing an outline is sufficient. 

Our experiments have no human reviewer, so any claims flagged as conflicts are withheld from the rewrite rather than resolved. Resolving a conflict means deciding which source to believe, and adjudicating source trust remains an open problem, so we withhold conflicts as a default.

%% file: figures/doc_diff.tex
\begin{figure}
    \centering
    \includegraphics[width=\linewidth]{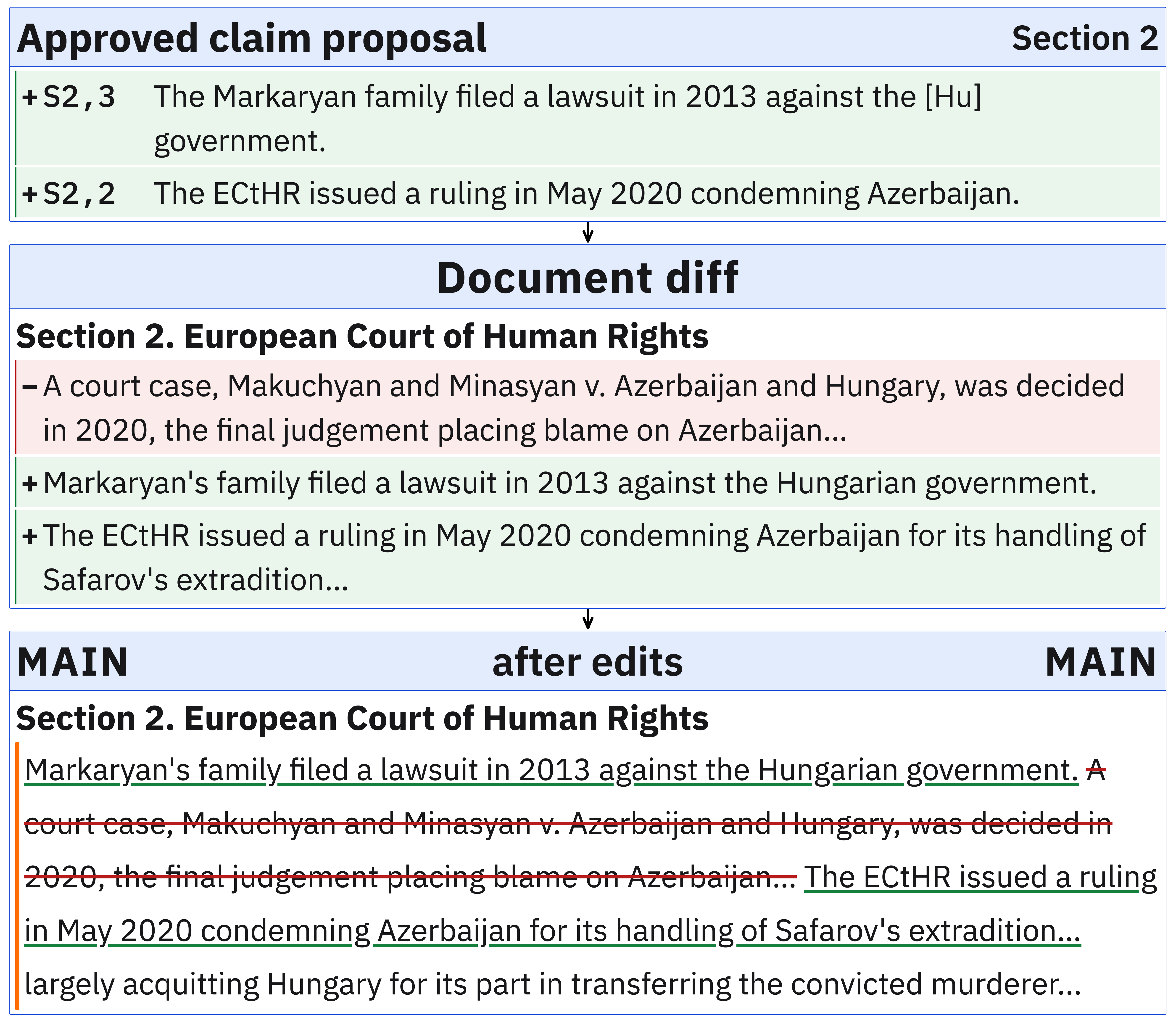}
    \caption{The approved claim proposal is integrated into each new and affected section, and diffing the result against the original gives the document diff. Sections with no proposed claims are left unchanged.}
    \label{fig:doc_diff}
    \vspace{-1em}
\end{figure}

%% file: sections/50-wiki-exps.tex
Our first application, cross-lingual knowledge revision, integrates knowledge from one language edition of Wikipedia into another. Events, entities, and locations are often documented unevenly across editions, covered more thoroughly in the language of the region they concern and sparsely elsewhere, depending on the distribution of volunteer editors. Each edition is therefore a source of human-authored knowledge that may be missing from or in conflict with another.

\paragraph{Data and Task.}
We treat English Wikipedia as the main and its counterparts in other languages as sources. Our task is thus to revise the English article with knowledge from the other sources while preserving its existing content. This tests whether a KPR can work from existing, curated text rather than newly surfaced information. Our documents come from MegaWika~2.0 \citep{barham2025megawika2comprehensivemultilingual}, whose collections \citep{barham2023megawikamillionsreportssources,barham2025megawika2comprehensivemultilingual} are built specifically for broad multilingual Wikipedia coverage with aligned articles across languages.

\paragraph{Evaluation.}
We evaluate these rewrites along three dimensions: the \emph{quality} of the rewritten article, the \emph{edit cost} of accepting it, and its value as a \emph{grounding source} for downstream question answering. \autoref{append:metric} gives more evaluation details.

\input{tables/main-table-performance}
\input{tables/wiki-mirage}

\emph{Quality.} 
We measure quality with MiRAGE~\citep{martin2026seeingmirageevaluatingmultimodal}. Information precision (\textbf{InfoP}) is a source-constrained variant of FActScore: each claim in the rewrite is verified against the documents used to produce it, checking that added claims faithfully reflect their sources rather than being distorted during rewriting. We adapt information recall (\textbf{InfoR}) to measure the coverage of two distinct claim sets: (1) \textit{InfoR-R}~(retain), the
original English claims preserved in the rewrite, and (2) \textit{InfoR-A}~(add), source claims that are added in the rewrite. These capture whether the rewrite preserves existing content and adds the intended new content, respectively.

\emph{Grounding Source.}
We test how well each rewritten article serves as a grounding source for question answering. To build the QA set, we take the decomposed claims from the English and multilingual articles, generate QA pairs from them, and filter for quality (answerable, not context-dependent, well-answered), yielding English and multilingual splits. With the two splits, we measure whether the rewrite \emph{loses} existing knowledge (En-QA) and whether it \emph{adds} the new cross-lingual knowledge (Multi-QA).

\emph{Edit Cost.}
We measure edit cost from the perspective of a reviewer who must review and approve a method's changes to a document. From a word-level diff between the main $M$ and the rewrite $R$, we report five quantities: word edit rate (\textbf{WER}), the word-level Levenshtein distance per source word; \textbf{Click}, the number of contiguous edit blocks the reviewer must approve, regardless of their size; added tokens (\textbf{Tok}), the number of new tokens generated; preservation (\textbf{Presv}), the fraction of the main untouched; and expansion (\textbf{Add}), the length ratio $\frac{R}{M}$. WER, Click, and Tok measure the cost of producing and reviewing a rewrite, while Presv and Add describe the shape of the rewrite.

We evaluate the QA accuracy of Qwen3.5-9B/27B~\citep[Q3.5-XB;][]{qwen35blog}, Qwen3-8B/30B~\citep[Q3-XB;][]{yang2025qwen3technicalreport}, Gemma-4-31B~\citep[G4-31B;][]{gemmateam2026gemma4technicalreport}, Llama-3.3-70B, Llama-3.1-8B, and Llama-4-Scout~\citep[L3.X-XB; L4-Scout;][]{grattafiori2024llama3herdmodels, meta2025llama4}, Mixtral-8x7B~\citep[M-8x7B;][]{jiang2024mixtralexperts}, OLMo-3-7B \citep[OLMo-3-7B;][]{olmo2026olmo3}, Nemotron 3 \cite[N3-120B;][]{nvidia2025nvidianemotron3efficient} and GPT Sol 5.6 \cite[Sol-5.6;][]{openai2026gpt56} when conditioning on each rewritten article.

\input{tables/small-models}

\subsection{Article Quality}
\autoref{tab:wiki-mirage} reports the InfoP and the two InfoR variants. We find that conditioning on claims rather than raw text (ConText vs. ConClaim) raises both the precision and the recall of added information. When adding the claim proposal (ConClaim vs. KPR), both precision and recall rise again, with the larger gain in added information. Retention is high for every method, so the methods differ mainly in how much of the source knowledge they integrate, where KPRs see the largest benefit.

\subsection{Grounding Source}
\autoref{tab:full-qa-results} reports each model's accuracy under five conditions: no document (closed-book), the original main (EW), and the main after each rewrite (ConText, ConClaim, and KPR).

\input{tables/frontier-models-wiki}

\paragraph{Multilingual QA.}
The unchanged article is a poor grounding source, averaging below even closed-book, because the queried facts are absent from the English article, and some models abstain rather than guess.\footnote{Gemma-4 and Qwen3.5 faithfully abstain on  around 70\%.} All three rewrites recover some of this missing knowledge, but the KPR method recovers substantially more than either baseline. This shows the downstream impact of KPR's high InfoR~(add), demonstrating KPRs incorporate multilingual information in a way that grounding articles can properly utilize. 

\paragraph{English QA.}
On questions about content already in the English article, the unchanged article scores highest (EW). All three rewrites degrade QA performance slightly, but stay within half a percent of one another, so integrating the cross-lingual knowledge does not meaningfully cost existing content for any method. KPR therefore obtains its multilingual gains without giving up English performance relative to the baselines.

\paragraph{Small Models.}
The KPR advantage does not require a large model to exploit the rewritten article. On smaller models (\autoref{tab:small-qa}), we still find that conditioning on the KPR-revised article provides a strong lift in performance. Small models running locally are exactly where a personal, grounded wiki is most useful.\footnote{As envisioned by Karpathy: \url{https://gist.github.com/karpathy/442a6bf555914893e9891c11519de94f}} A KPR gives such a setup a document worth grounding on, without a frontier model or a search service in the loop.

\paragraph{KPRs vs.\ Search.}
In \autoref{tab:frontier-qa}, we test whether grounding on a KPR is worth it over a frontier model with search. We create a QA set of the hardest questions by taking a random sample of 100 multilingual QA instances that no dense model could answer. On this set, we again see that grounding on KPR recovers the most answers at every scale.\footnote{We provide baseline conditioned results in \autoref{append:wiki-details}.} We additionally give Sol-5.6 access to web search, and it recovers only marginally more answers than EW. Because the questions are posed in English, while the answer may only be documented in its source language, we translate each question to its source language for closed book (CB-T) and web search (WS-T). This helps both, but neither approaches KPR, and even the smallest 7B model grounded on the KPR article outperforms the frontier model with search in either condition.

The sources are other-language Wikipedia editions, so the knowledge is public and indexed, but not surfaced by search. Having access to a KPR-revised document provides a more reliable grounding source than relying on search at query time.

\input{tables/wiki-edit-cost}

\subsection{Edit Cost}
\autoref{tab:wiki-edit} reports the cost of authoring and reviewing the revisions and their shape. We find that KPRs preserve the most of the article and make changes in the smallest number of contiguous edits. While both ConText and KPRs add a similar number of new tokens (2.3$\times$ context increase), KPRs integrate substantially more of the source knowledge and require a reviewer looking at the text to approve fewer separate changes. ConClaim makes the smallest change to the article overall, but requires more than twice the approvals of a KPR. Without a claim proposal, every claim routed to a section is passed to the rewrite, so the model revises content that did not need to change.

%% file: tables/main-table-performance.tex
\begin{table*}
    \centering
    \begin{tabular}{lc|cc|ccc}
    \toprule
        \textbf{Language} & \textbf{Model} & \textbf{CB} & \textbf{EW} & \textbf{ConText} & \textbf{ConClaim} & \textbf{KPR} \\
    \midrule
        \multirow{7}{*}{En-QA}
            & Q3.5-27B & 33.6 & 95.7 & 91.5 & 91.5 & 90.9 \\
            & Q3-30B   & 27.7 & 92.3 & 88.3 & 88.7 & 88.1 \\
            & G4-31B   & 33.3 & 95.1 & 91.1 & 91.0 & 90.2 \\
            & L3.3-70B & 45.4 & 95.1 & 91.9 & 92.1 & 91.6 \\
            & L4-Scout & 32.6 & 77.2 & 67.3 & 70.1 & 71.0 \\
            & M-8x7B   & 36.1 & 93.3 & 90.4 & 90.2 & 89.7 \\
            & N3-120B  & 42.1 & 95.3 & 92.5 & 92.6 & 92.2 \\
        \midrule
            & Avg      & 35.8 & 92.0 & 87.6 & \textbf{88.0} & 87.7 \\
    \midrule
        \multirow{7}{*}{Multi-QA}
            & Q3.5-27B & 36.3 & 15.5 & 41.7 & 54.6 & 67.8 \\
            & Q3-30B   & 31.0 & 29.4 & 49.1 & 59.9 & 70.7 \\
            & G4-31B   & 35.9 & 14.6 & 39.5 & 52.7 & 66.3 \\
            & L3.3-70B & 43.5 & 31.9 & 52.9 & 62.8 & 74.0 \\
            & L4-Scout & 33.9 & 34.5 & 42.3 & 48.9 & 54.8 \\
            & M-8x7B   & 36.4 & 39.0 & 55.7 & 64.2 & 74.5 \\
            & N3-120B  & 41.6 & 41.3 & 57.1 & 66.1 & 76.6 \\
        \midrule
            & Avg      & 36.9 & 29.5 & 48.4 & 58.5 & \textbf{69.2} \\
    \bottomrule
    \end{tabular}
    \caption{QA accuracy, conditioned on the original English article (EW), each rewrite (ConText, ConClaim, KPR), or no document (CB). Q: Qwen, G: Gemma, L: Llama, M: Mixtral, N: Nemotron.}
    \label{tab:full-qa-results}
    \vspace{-1em}
\end{table*}

%% file: tables/wiki-mirage.tex
\begin{table}[t]
    \centering
    \setlength{\tabcolsep}{4pt}
    \begin{tabular}{l|ccc}
    \toprule
        \textbf{Method} & \textbf{InfoP} & \textbf{InfoR-R} & \textbf{InfoR-A} \\
    \midrule
        ConText   & 0.837 & 0.946 & 0.716 \\
        ConClaim  & 0.853 & 0.943 & 0.786 \\
        KPR       & \textbf{0.878} & \textbf{0.950} & \textbf{0.891} \\
    \bottomrule
    \end{tabular}
    \caption{Article-quality evaluation of the Wikipedia rewrites, measured with MiRAGE. InfoP: information precision; InfoR-R: Information Recall \textbf{R}etained from the English article; InfoR-A: Information Recall \textbf{A}dded by the sources.}
    \label{tab:wiki-mirage}
    \vspace{-1em}
\end{table}

%% file: tables/small-models.tex
\begin{table}
    \centering
    \begin{tabular}{c|cc|c}
    \toprule
        \textbf{Model} & \textbf{CB} & \textbf{EW} & \textbf{KPR} \\
    \midrule
        Qwen3.5-9B   & 37.4 & 15.5 & \textbf{44.8} \\
        Qwen3-8B     & 33.5 & 35.8 & \textbf{61.4} \\
        Llama-3.1-8B & 38.9 & 33.8 & \textbf{61.5} \\
        OLMo-3-7B    & 25.5 & 25.1 & \textbf{51.1} \\
    \bottomrule
    \end{tabular}
    \caption{Smaller models on multilingual QA.}
    \label{tab:small-qa}
    \vspace{-1em}
\end{table}

%% file: tables/frontier-models-wiki.tex
\begin{table}
    \centering
    \setlength{\tabcolsep}{3pt}
    \begin{tabular}{c|cc|c|c|cc}
    \toprule
        \textbf{Model} & \textbf{CB} & \textbf{CB-T} & \textbf{EW} & \textbf{KPR} & \textbf{WS} & \textbf{WS-T}\\
    \midrule
        \rowcolor{gray!15} Q3.5-9B  & 2 & 3 & 2  & \textbf{49} & $-$ & $-$ \\
        \rowcolor{gray!15} Q3-8B    & 4 & 4 & 9  & \textbf{56} & $-$ & $-$ \\
        \rowcolor{gray!15} L3.1-8B  & 2 & 1 & 7  & \textbf{56} & $-$ & $-$ \\
        \rowcolor{gray!15} O3-7B    & 2 & 2 & 13 & \textbf{50} & $-$ & $-$ \\
    \midrule
        \rowcolor{gray!15} Q3.5-27B & 0 & 5 & 3  & \textbf{52} & $-$ & $-$ \\
        \rowcolor{gray!15} Q3-30B   & 0 & 3 & 12 & \textbf{54} & $-$ & $-$ \\
        \rowcolor{gray!15} G4-31B   & 0 & 4 & 4  & \textbf{54} & $-$ & $-$ \\
        \rowcolor{gray!15} L3.3-70B & 0 & 5 & 8  & \textbf{51} & $-$ & $-$ \\
    \midrule
        Sol-5.6   & 13 & 20 & 32 & \textbf{62} & 38 & 41 \\
    \bottomrule
    \end{tabular}
    \caption{Model performance on the hardest QA set: 100 multilingual questions no dense open model answers closed-book. CB: closed-book; CB-T: closed book translated; EW: original English article; KPR: KPR-revised article; WS: web search; WS-T: web search translated.}
    \label{tab:frontier-qa}
    \vspace{-1em}
\end{table}

%% file: tables/wiki-edit-cost.tex
\begin{table}[t]
    \centering
    \setlength{\tabcolsep}{4pt}
    \begin{tabular}{l|c|cccc}
    \toprule
        \textbf{Method} & \textbf{WER} & \textbf{Click} & \textbf{Tok} & \textbf{Presv} & \textbf{Add} \\
    \midrule
        KPR & 131 & 11.2 & 1{,}299 & 97.3 & 2.3 \\
        ConClaim & 92 & 25.3 & 898 & 94.6 & 1.8 \\
        ConText & 138 & 14.7 & 1{,}374 & 96.9 & 2.3 \\
    \bottomrule
    \end{tabular}
    \caption{Cost of accepting the rewrite from the original Wikipedia article. WER: word edit rate, Click: number of contiguous edits a reviewer must approve, Tok: added tokens, Presv: percentage of the original document preserved in the rewrite, Add: length of the rewrite relative to the original document.}
    \label{tab:wiki-edit}
    \vspace{-1em}
\end{table}

%% file: sections/60-ragtime-exps.tex
Our second application moves from a single evolving article to \emph{query-driven} report generation, where the document is a report synthesized to answer a query and the sources are documents retrieved for it. This setting is harder than the Wikipedia revision in two ways. First, the sources are no longer aligned versions of the same article. Instead, they are independent documents that may overlap with, extend, or contradict one another and the existing report. Second, reports are regenerated as new sources surface---a query asked today may be re-answered next month against updated evidence---which is precisely where query-driven systems are weakest. Current deep research and report-generation systems regenerate from scratch~\cite{google2025geminidr}, discarding prior content as wasted compute. A KPR instead treats each regeneration as an incremental update to the standing report.

\input{tables/ragtime-mirage}

\paragraph{Data and Task.}
We build our experiments on RAGTIME \citep{lawrie2026overviewtrec2025ragtime}, a multilingual report-generation task in which a system is given a persona and query and performs retrieval-augmented generation over a collection of multilingual documents.\footnote{We do not perform retrieval; we use the relevance judgments to obtain the source documents directly.} From RAGTIME we construct three two-round variants of the task and report their details in \autoref{append:ragtime-details}.

\emph{Temporal}: new information arrives over time. RAGTIME metadata contains the document date, so we set a per-topic knowledge cutoff such that half the relevant documents surface only in round~2, simulating the same query asked at two times.

\emph{Conflict}: a round-2 source contradicts round-1 content. RAGTIME nuggets, QA evaluation units similar to atomic facts \citep{voorhees2003overview}, sometimes take different values across documents (e.g., a margin of victory reported as 2.7\% in one and 3\% in another). We construct these rounds so that conflicting nuggets appear in each.

\emph{Balanced}: information is split evenly, with an equal number of nuggets appearing in round-1 and round-2, maximizing how much a method must add while still preserving round-1.

\paragraph{Evaluation.}
To generate the round-1 report, we condition on the text the way a RAG system might put the relevant documents in context with the query to write the report. We evaluate report quality with MiRAGE~\citep{martin2026seeingmirageevaluatingmultimodal}, which has higher human agreement than AutoArgue \citep{walden2026autoarguellmbasedreportgeneration} on this task, using the same information metrics from the Wikipedia experiments. InfoP measures precision against the sources used in generation, InfoR-R (retain) the round-1 nuggets preserved in the updated report, and InfoR-A (add) the round-2 nuggets newly incorporated. We omit ConClaim in this setting, as the unfiltered claim sets from RAGTIME's source documents exceed the context window.

\input{tables/ragtime-edit-cost}

\subsection{Report Quality}
\autoref{tab:ragtime-mirage} reports the InfoP and the two InfoR variants across all three settings. KPR achieves the highest recall of both round-1 and round-2 nuggets in every setting, and leads on precision among the rewrites. Scratch adds the least information between rounds, while ConText has the lowest precision of any method in every setting. The round-1 report scores highest on precision, only needing to be faithful to the round-1 sources.

Every method recalls more round-1 nuggets than the round-1 report. The incremental methods never revisit the round-1 documents, but some round-1 information recurs in round-2, so integrating those sources recovers nuggets the original missed. 

\paragraph{Conflict.} 
When sources disagree across rounds, ConText integrates less of the round-2 information than it did in either other setting. Without a mechanism to flag conflicts it must resolve them implicitly during the rewrite, and doing so appears to suppress how much of the incoming information it incorporates. KPR instead withholds conflicting claims, adding new facts at a rate comparable to the other settings while holding the highest precision of any method that integrates a meaningful amount.

\subsection{Edit Cost}
\autoref{tab:rt-edit} reports the cost of authoring and reviewing each update. We find that Scratch's high precision makes sense in the context of its edit shape, generating short, but high-confidence reports. This also makes it the most expensive to review, as only coincidental n-gram overlaps are preserved in the generation. ConText and KPR instead make similarly sized updates to the round-1 document, but ConText struggles to add as much information to the updates as KPR.

%% file: tables/ragtime-mirage.tex
\begin{table}[t]
    \centering
    \setlength{\tabcolsep}{4.5pt}
    \begin{tabular}{ll|ccc}
    \toprule
        & \textbf{Round-2} & \textbf{InfoP} & \textbf{InfoR-R} & \textbf{InfoR-A} \\
    \midrule
        \multirow{4}{*}{Temp.}
            & \cellcolor{gray!15}R1 Main & \cellcolor{gray!15}\textbf{0.928} & \cellcolor{gray!15}0.416 & \cellcolor{gray!15}$-$ \\
            & Scratch & 0.882 & 0.475 & 0.223 \\
            & ConText & 0.620 & 0.649 & 0.632 \\
            & KPR     & 0.729 & \textbf{0.709} & \textbf{0.682} \\
    \midrule
        \multirow{4}{*}{Conf.}
            & \cellcolor{gray!15}R1 Main & \cellcolor{gray!15}\textbf{0.904} & \cellcolor{gray!15}0.430 & \cellcolor{gray!15}$-$ \\
            & Scratch & 0.875 & 0.456 & 0.286 \\
            & ConText & 0.666 & 0.625 & 0.429 \\
            & KPR     & 0.811 & \textbf{0.782} & \textbf{0.571} \\
    \midrule
        \multirow{4}{*}{Bal.}
            & \cellcolor{gray!15}R1 Main & \cellcolor{gray!15}\textbf{0.961} & \cellcolor{gray!15}0.453 & \cellcolor{gray!15}$-$ \\
            & Scratch & 0.862 & 0.558 & 0.326 \\
            & ConText & 0.692 & 0.739 & 0.537 \\
            & KPR     & 0.841 & \textbf{0.856} & \textbf{0.632} \\
    \bottomrule
    \end{tabular}
    \caption{RAGTIME results across each setting. InfoR-R: information recall retained from the round-1 report; InfoR-A: information recall added from the round-2 sources. For the round-1 report (R1 Main), InfoR-R is the recall of round-1 nuggets in the original report.}
    \label{tab:ragtime-mirage}
    \vspace{-1em}
\end{table}

%% file: tables/ragtime-edit-cost.tex
\begin{table}[t]
    \centering
    \setlength{\tabcolsep}{4pt}
    \begin{tabular}{l|c|cccc}
    \toprule
        \textbf{Method} & \textbf{WER} & \textbf{Click} & \textbf{Tok} & \textbf{Presv} & \textbf{Add} \\
    \midrule
        Scratch & 130 & 56.1 & 770 & 37.3 & 1.2 \\
        ConText & 330 & 12.5 & 3{,}091 & 98.2 & 4.3 \\
        KPR & 372 & 20.5 & 3{,}475 & 96.3 & 4.7 \\
    \bottomrule
    \end{tabular}
    \caption{Cost of accepting the rewrite from Round-1 RAGTIME document. WER: word edit rate, Click: number of contiguous edits a reviewer must approve, Tok: added tokens, Presv: percentage of the original document preserved in the rewrite, Add: length of the rewrite relative to the original document.}
    \label{tab:rt-edit}
    \vspace{-1em}
\end{table}

%% file: sections/70-findings.tex
\paragraph{Claims as the unit for information.}
We demonstrate that claims are a better unit of information to operate over than text when integrating knowledge. In both experiments, conditioning the rewrite on claims rather than source text improves both the precision and the recall of added information, with the same trend showing in the QA experiments. Claim-level representations are known to be the appropriate unit for factuality evaluation \citep{min-etal-2023-factscore, song-etal-2024-veriscore} and can even boost retrieval performance \citep{chen-etal-2024-dense}. \emph{Our results extend this finding to knowledge integration, where conditioning generation on claims rather than source text yields a more faithful grounding document.}

\paragraph{Claim proposals improve the rewrite.}
We find that claim proposals consistently improve the quality of the rewrite over the baselines, maintaining precision and raising the recall of integrated information. The proposal also helps shape the rewrite into contiguous, easy to review edits. A KPR also adds the most information per token generated, as baselines writing a comparable volume of text integrate substantially less of the source information.

%% file: sections/80-conclusion.tex
We introduce Knowledge Pull Requests, which reframe continual document authoring as a reviewable, knowledge-level operation rather than an uninterpretable rewrite. By decomposing sources into claims, filtering and routing them, and surfacing conflicts, a KPR produces a ChangeLog that separates what knowledge changes from how the text changes. Across cross-lingual Wikipedia revision and query-driven report updating, KPRs integrate new information more completely and preserve existing content better than other methods, while concentrating their changes into contiguous edits a reviewer can approve. By making the unit of change a reviewable claim rather than an opaque edit, KPRs open a path toward collaborative, human-in-the-loop authoring.

\section*{Limitations}
\paragraph{Computational Efficiency.}
Our pipeline performs every step---decomposition, classification, routing, and rewriting---with an LLM. While a KPR rewrites only the sections that change rather than regenerating the entire document, each intermediate step still requires an LLM call, and cheaper alternatives exist. For example, a lightweight encoder could classify claim containment or route claims to sections in place of prompting. Our edit-cost metrics measure the size and shape of the resulting rewrite, not the compute spent producing it. Reducing this per-step cost is important for deploying KPRs at scale (e.g., continually updating Wikipedia) and is left to future work.d

\paragraph{Human Review.}
A central motivation for KPRs is that the ChangeLog is inspectable and reviewable. A human can assess proposed claims, resolve flagged conflicts, and approve or reject changes before they are merged, analogous to a code review. However, our experiments run the pipeline automatically, withholding flagged conflicts from the rewrite, and thus do not evaluate the review process itself. Human studies that investigate whether a ChangeLog makes an editor faster and more accurate than a text diff, and how claim proposals should be adjudicated by a human, are both necessary future work.

%% file: appendix/wiki-details.tex
\paragraph{Articles.}
We sample 599 articles from MegaWika~2.0 \citep{barham2025megawika2comprehensivemultilingual},
selecting evenly from the distribution of cross-lingual links per document so that the source language editions and number of source articles vary across topics. We cover 49 of MegaWika's 50 languages~(all but English, our target): af, ar, az, bn, cs, de, es, et, fa, fi, fr, ga, gl, gu, he, hi, hr, id, it, ja, ka, kk, km, ko, lt, lv, mk, ml, mn, mr, my, ne, nl, pl, ps, pt, ro, ru, si, sl, sv, ta, th, tr, uk, ur, vi, xh, zh.

\paragraph{QA Generation.}
To build the question-answering set that measures grounding, we convert every claim in an article into a question with the claim as its gold answer. For the multilingual split, we additionally discard claims that are contained by English Wikipedia, so that the remaining questions target knowledge genuinely available only in other language editions. We then filter the questions for quality and deduplicate them, keeping only questions that (i) are specific, well-formed factoid questions with a clear, verifiable answer; (ii) are answerable on their own, without external context (e.g., ``who is the captain of the ship?'' is discarded, as it depends on an unstated referent); and (iii) are cleanly and correctly answered by the gold answer, taking the claim as ground truth. \autoref{tab:qa-generation} reports the number of questions remaining after each stage. We provide our prompts for filtering the data in \autoref{prompt:gold-audit}, \autoref{prompt:question-audit}, and \autoref{prompt:claim-review}.

\paragraph{100-Question Subset.}
To construct this set, we filter the multilingual QA questions down to those not answered correctly by any of our dense models in a closed-book, non translated setting, and then sample 100 at random. This isolates questions whose answers are absent from the models' parametric knowledge, so that any performance must come from the grounding document. For the translated versions of this subset (CB-T, WS-T), we translate questions with Qwen3.5-27B.

\paragraph{Frontier Model Usage.}
We access GPT-5.6 (Sol) through the OpenAI API, setting reasoning effort to its maximum and retaining this setting in the web-search condition. In \autoref{tab:frontier-qa-extra} we also show performance when conditioning on ConText and ConClaim.

\input{tables/appendix-qa-stats}

\input{tables/frontier-qa-extra}

%% file: tables/appendix-qa-stats.tex
\begin{table}[]
    \centering
    \begin{tabular}{l|c|c}
    \toprule
        \textbf{Stage} & \textbf{Multilingual} & \textbf{English} \\
    \midrule
        Original & 427,898 & 111,693 \\
        Absent/Conflict & 81,490 & $-$ \\
        Final & 44,683 & 26,572 \\
    \bottomrule
    \end{tabular}
    \caption{Question counts at each stage of QA generation. Original: questions generated from all article claims. Absent/Conflict: multilingual questions retained after keeping only claims absent from or contradicting English Wikipedia. Final: questions remaining after quality filtering and deduplication.}
    \label{tab:qa-generation}
\end{table}

%% file: tables/frontier-qa-extra.tex
\begin{table*}
    \centering
    \begin{tabular}{c|cc|c|cc|c|cc}
    \toprule
        Model & CB & CB-T & English-Wiki & ConText & ConClaim & KPR & WS & WS-T\\
    \midrule
        \rowcolor{gray!15}
        Q3.5-27B & 0 & 5 & 3  & 18 & 31 & \textbf{52} & $-$ & $-$ \\
        \rowcolor{gray!15}
        Q3-30B   & 0 & 3 & 12 & 26 & 34 & \textbf{54} & $-$ & $-$ \\
        \rowcolor{gray!15}
        G4-31B   & 0 & 4 & 4  & 20 & 32 & \textbf{54} & $-$ & $-$ \\
        \rowcolor{gray!15}
        L3.3-70B & 0 & 5 & 8  & 25 & 31 & \textbf{51} & $-$ & $-$ \\
    \midrule
        Sol-5.6  & 13 & 20 & 32 & 46 & 54 & \textbf{62} & 38 & 41 \\
    \bottomrule
    \end{tabular}
    \caption{Frontier and open model performance on the hardest QA set (\autoref{tab:frontier-qa}), including the ConText and ConClaim rewrite conditions omitted from the main table. CB-T: closed-book translated; WS-T: web search translated.}
    \label{tab:frontier-qa-extra}
\end{table*}

%% file: appendix/ragtime-details.tex
\input{tables/ragtime-nugget-splits}

\paragraph{Split Creation.}
We construct all three splits programmatically from the RAGTIME relevance judgments, reporting the resulting \textbf{novel} nugget counts per round in \autoref{tab:ragtime-splits}.
\begin{itemize}
    \item \textbf{Temporal}: we choose a per-topic knowledge cutoff date that divides the relevant documents into two equal halves by publication date, assigning the earlier half to round~1 and the later half to round~2.
    \item \textbf{Conflict}: RAGTIME distinguishes \emph{OR} nuggets, which admit multiple acceptable answers across documents, from \emph{AND} nuggets, whose supporting information is spread across documents. We place the documents supporting an OR nugget's conflicting answers (or an AND nugget's complementary pieces) in different rounds, so that a conflicting or complementary document surfaces in round~2.
    \item \textbf{Balanced}: we partition documents to make the number of nuggets appearing in each round as equal as possible.
\end{itemize}

%% file: tables/ragtime-nugget-splits.tex
\begin{table}[]
    \centering
    \begin{tabular}{l|cc}
    \toprule
        \textbf{Setting} & \textbf{Round-1} & \textbf{Round-2} \\
    \midrule
        Temporal        & 12.2 & 3.2 \\
        Conflict & 14.7 & 0.5 \\
        Balanced            & 7.6  & 8.0 \\
    \bottomrule
    \end{tabular}
    \caption{Novel nuggets introduced per round in each RAGTIME variant (nuggets not present in the prior round).}
    \label{tab:ragtime-splits}
\end{table}

%% file: appendix/method-details.tex
\input{tables/appendix-decomp-numbers}
\paragraph{Claim Decomposition.}
Claim decomposition is less straightforward across languages than within a single one. A claim must end up as an atomic, decontextualized statement in the target language, faithful to its meaning in the source language and without propagating translation errors. We compare three strategies (\autoref{tab:decomp-stats}):
\begin{itemize}
    \item \textbf{Translate-then-decompose}: translate the source sentence into the target language, then decompose claims from the translation. We use both the MegaWika NLLB translations \cite{nllbteam2022languageleftbehindscaling} and Qwen3.5-9B.
    \item \textbf{Native-then-translate}: decompose the sentence into claims in the source language, then translate those claims to the target language.
    \item \textbf{Cross-lingual decomposition}: decompose the source sentence directly into claims in English.\footnote{Note that the English-to-English decomposition is there as an example baseline to compare the cross-lingual decomposition against.}
\end{itemize}
Translate-then-decompose is the least faithful under either translator, as translation errors compound with decomposition. Native-then-translate and cross-lingual decomposition achieve roughly the same faithfulness. We use cross-lingual decomposition because it uses a single LLM pass instead of two passes.

\paragraph{Prompts for each method.}
The KPR prompts are given in \autoref{prompt:claim-review} (claim review), \autoref{prompt:route-claims} (routing), \autoref{prompt:route-new-sections} (new-section routing), \autoref{prompt:rewrite} (rewriting), and \autoref{prompt:rewrite-new-section} (new-section rewriting). ConClaim and ConText reuse the same routing and rewriting flow as KPR and differ from KPR only as follows:
\begin{itemize}
    \item \textbf{ConClaim} omits the claim-review step (\autoref{prompt:claim-review}), so claims are not filtered by absent/covered/conflicting/relevance before rewriting.
    \item \textbf{ConText} conditions each section's rewrite on raw source text, with an LLM classifying section relevance in place of claim review and routing. Its rewriting prompts are reworded to take source text as input, which we omit, as they are nearly identical to the KPR versions
\end{itemize}

%% file: tables/appendix-decomp-numbers.tex
\begin{table*}[ht]
    \centering
    \begin{tabular}{lc|cccc}
    \toprule
        \textbf{Strategy} & \textbf{Claims} & \textbf{Supported} & \textbf{Contradicted} & \textbf{Hallucinated} & \textbf{Cannot-Det} \\
    \midrule
        English-to-English & 175{,}267 & 88.7 & 0.7  & 10.0 & 0.7 \\
    \midrule
        NLLB Translate & 431{,}098 & 79.8 & 2.3  & 16.8 & 1.1 \\
        LLM Translate & 451{,}333 & 84.4 & 1.0  & 13.2 & 1.4 \\
        Native Decomp & 376{,}331 & \textbf{89.8} & 1.1  &  \textbf{8.3} & \textbf{0.8} \\
        Cross Decomp & 357{,}460 & 89.0 & \textbf{0.9}  &  9.2 & 0.9 \\
    \bottomrule
    \end{tabular}
    \caption{Claim decomposition strategies evaluated by verdict distribution. Each row shows the number of claims produced and the percentage judged supported, contradicted, hallucinated, or cannot-determine when verified against the source.}
    \label{tab:decomp-stats}
\end{table*}

%% file: appendix/metric-details.tex
\subsection{Quality}
We measure quality with MiRAGE~\citep{martin2026seeingmirageevaluatingmultimodal}, which scores a predicted text $P$ against a set of reference evidence $R$ by decomposing both into atomic subclaims and measuring support with a scoring function $\textsf{s}(\cdot,\cdot)\in[0,1]$. In every experiment $\textsf{s}$ is a cross-lingual LLM support judge (Qwen3.5-27B). Following the definitions, we report Information Precision (InfoP), and two variations of Information Recall (InfoR-A, InfoR-R).

\subsubsection{Information Precision}
InfoP is a source-constrained variant of FActScore~\cite{min-etal-2023-factscore}. For each $P$, we decompose the sentences into subclaims $C_P$ and score each for support against $R$:
\begin{equation}
    \mathrm{InfoP}(C_P,R)=\frac{1}{|C_P|}\sum_{c\in C_P}\textsf{s}(c,R).
\end{equation}
We use the Collection variant of InfoP, where the documents are used, instead of their underlying claims. For Wikipedia, $P$ is the rewrite's added sentences and $R$ is the non-English sources. We do not do the full articles here because it is not expected for the unmodified English Article to be supported by the non-English sources. For RAGTIME, $P$ is the round-2 report and $R$ is the union of round-1 and round-2 source documents. 

\subsubsection{Information Recall}
InfoR follows the same formulation, with $P$ and $R$ reverse, scoring claims decomposed from the reference $C_R$ against $P$.
\begin{equation}
    \mathrm{InfoR}(C_R,P)=\frac{1}{|C_R|}\sum_{c\in C_R}\textsf{s}(c,P).
\end{equation}
For both tasks, we introduce two variants of InfoR: retain (InfoR-R), the number of claims preserved in the rewrite, and add (InfoR-A), the number of source claims added in the rewrite. For Wikipedia, InfoR-R has $C_R$ be the set of decomposed claims of the original English article, measuring how much of the original English Wikipedia the rewrite preserves; and InfoR-A has $C_R$ be the set of decomposed claims of the non-English sources, measuring how much was added from the non-English sources during rewriting. For RAGTIME, InfoR-R has $C_R$ be the set of nuggets from round-1, measuring how much of the round-1 report the rewrite preserves; and InfoR-A has $C_R$ be the set of nuggets from round-2, measuring how much was added from the new sources during the rewriting.

\subsection{Edit Cost}
Each metric compares the original token sequence $M$ to the rewritten one $R$ via the word-level diff, which segments the change into maximal contiguous blocks: \emph{equal}, \emph{insert}, \emph{replace}, \emph{delete}. Ratio metrics are token-weighted over the document set $\mathcal{D}$ to not overweight short documents. 

\begin{itemize}
  \item \textbf{Word edit rate (WER)}: the word-level Levenshtein distance (insertions, deletions, substitutions) between $M$ and $R$,
  \[
    \mathrm{WER}=\frac{\sum_{d\in\mathcal{D}}\mathrm{edits}(M_d,R_d)}{\sum_{d\in\mathcal{D}}|M_d|}\times 100\%,
  \]
  it can exceed $100\%$ when a rewrite adds more than the original length.
  
  \item \textbf{Click}: one click of the ``approve'' button per maximal contiguous \emph{inserted} or \emph{replaced} span, regardless of its length,
  \[
    \mathrm{Click}=\big|\{\text{\emph{insert} and \emph{replace} blocks}\}\big|,
  \]
  reported as the per-document mean. Many scattered small edits cost more clicks than a few large contiguous blocks, even when the latter add more total text.
  
  \item \textbf{Added tokens (Tok)}: the number of new word tokens,
  \[
    \mathrm{Tok}=\!\!\sum_{\text{\emph{insert},\,\emph{replace} blocks}}\!\!|R\text{-span}|,
  \]
  the per-document mean of the rewrites token count.
  
  \item \textbf{Preservation \% (Presv)}: the percentage of the original document the method keeps verbatim, as contiguous \emph{equal} spans,
  \[
    \mathrm{Presv}=\frac{\sum_{d}\sum_{\text{\emph{equal} blocks}}|M\text{-span}|}{\sum_{d}|M_d|}\times 100\%.
  \]
  High preservation means the method left most of the original in place; low means it regenerated the content.
  \item \textbf{Add} (expansion): how many times larger the rewritten document is than the original,
  \[
    \mathrm{Add}=\frac{\sum_{d}|R_d|}{\sum_{d}|M_d|}
  \]
\end{itemize}
For the Wikipedia revisions $M$ is the English article and $R$ the reconstructed rewrite. For RAGTIME $M$ is the round-1 seed report and $R$ the round-2 report. WER, Click, and Tok measure the cost of producing and reviewing a rewrite, while Presv and Add describe its shape. Note that a good revision may legitimately rewrite much of a document, so neither is better in a fixed direction.

%% file: appendix/prompts.tex
\begin{figure*}[t]
\begin{lstlisting}[style=prompt]
Instructions:
- You are given a paragraph and one sentence from the paragraph to decompose
- The text may be in any language — decompose the sentence into atomic claims
- Output the claims in the SAME LANGUAGE as the input (do not translate)
- You must output a JSON array: [{"claim": "..."}, {"claim": "..."}, ...]

##PARAGRAPH##: On 15 April 2019, just before 18:20 CEST, a structural fire broke out in the roof space of Notre-Dame de Paris, a medieval Catholic cathedral in Paris, France. By the time the fire was extinguished, the cathedral's wooden spire had collapsed, most of the wooden roof had been destroyed, and the cathedral's upper walls were severely damaged.
##SENTENCE##: On 15 April 2019, just before 18:20 CEST, a structural fire broke out in the roof space of Notre-Dame de Paris, a medieval Catholic cathedral in Paris, France.
##DECOMPOSITION##:
```json
[
    {"claim": "A structural fire broke out"},
    {"claim": "The fire broke out on 15 April 2019"},
    {"claim": "The fire broke out just before 18:20 CEST"},
    {"claim": "The fire broke out in the roof space"},
    {"claim": "Notre-Dame de Paris is a medieval Catholic cathedral"},
    {"claim": "Notre-Dame de Paris is located in Paris, France"}
]
```

##PARAGRAPH##: Le 15 avril 2019, peu avant 18h20 CEST, un incendie s'est déclaré dans la charpente de la cathédrale Notre-Dame de Paris, une cathédrale catholique médiévale située à Paris, en France. Au moment où l'incendie a été éteint, la flèche en bois de la cathédrale s'était effondrée, la majeure partie de la toiture en bois avait été détruite et les murs supérieurs de la cathédrale avaient été gravement endommagés.
##SENTENCE##: Le 15 avril 2019, peu avant 18h20 CEST, un incendie s'est déclaré dans la charpente de la cathédrale Notre-Dame de Paris, une cathédrale catholique médiévale située à Paris, en France.
##DECOMPOSITION##:
```json
[
    {"claim": "Un incendie s'est déclaré"},
    {"claim": "L'incendie s'est déclaré le 15 avril 2019"},
    {"claim": "L'incendie s'est déclaré peu avant 18h20 CEST"},
    {"claim": "L'incendie s'est déclaré dans la charpente"},
    {"claim": "Notre-Dame de Paris est une cathédrale catholique médiévale"},
    {"claim": "Notre-Dame de Paris est située à Paris, en France"}
]
```

##PARAGRAPH## [paragraph]
##SENTENCE## [sentence]
##DECOMPOSITION##:
\end{lstlisting}
\caption{\textbf{Native-then-translate}. Decomposes a sentence into atomic claims in the \emph{source} language.}
\label{prompt:native-then-translate}
\end{figure*}

\begin{figure*}[t]
\begin{lstlisting}[style=prompt]
Instructions:
- You are given a JSON array of claims written in a non-English language
- Translate each claim into English, preserving meaning exactly
- Output a JSON array in the same format: [{"claim": "..."}, {"claim": "..."}, ...]
- Do not add, remove, or merge claims — one-to-one translation only

##CLAIMS##: [{"claim": "Un incendie s'est déclaré"}, {"claim": "L'incendie s'est déclaré le 15 avril 2019"}, {"claim": "Notre-Dame de Paris est une cathédrale catholique médiévale"}, {"claim": "Notre-Dame de Paris est située à Paris, en France"}]
##TRANSLATION##:
```json
[
    {"claim": "A fire broke out"},
    {"claim": "The fire broke out on 15 April 2019"},
    {"claim": "Notre-Dame de Paris is a medieval Catholic cathedral"},
    {"claim": "Notre-Dame de Paris is located in Paris, France"}
]
```

##CLAIMS##: [claims]
##TRANSLATION##:
\end{lstlisting}
\caption{\textbf{Claim-translation prompt.} Translates
native-language claims into English one-to-one, without adding, removing or merging claims.}
\label{prompt:translate-claims}
\end{figure*}

\begin{figure*}[t]
\begin{lstlisting}[style=prompt]
Instructions:
- You are given a paragraph and one sentence from the paragraph to decompose
- The text may be in any language
- Decompose the sentence into atomic claims and output them in ENGLISH regardless of the input language
- You must output a JSON array: [{"claim": "..."}, {"claim": "..."}, ...]

##PARAGRAPH##: Le 15 avril 2019, peu avant 18h20 CEST, un incendie s'est déclaré dans la charpente de la cathédrale Notre-Dame de Paris, une cathédrale catholique médiévale située à Paris, en France. Au moment où l'incendie a été éteint, la flèche en bois de la cathédrale s'était effondrée, la majeure partie de la toiture en bois avait été détruite et les murs supérieurs de la cathédrale avaient été gravement endommagés.
##SENTENCE##: Le 15 avril 2019, peu avant 18h20 CEST, un incendie s'est déclaré dans la charpente de la cathédrale Notre-Dame de Paris, une cathédrale catholique médiévale située à Paris, en France.
##DECOMPOSITION##:
```json
[
    {"claim": "A structural fire broke out"},
    {"claim": "The fire broke out on 15 April 2019"},
    {"claim": "The fire broke out just before 18:20 CEST"},
    {"claim": "The fire broke out in the roof space"},
    {"claim": "Notre-Dame de Paris is a medieval Catholic cathedral"},
    {"claim": "Notre-Dame de Paris is located in Paris, France"}
]
```

##PARAGRAPH##: El huracán Irma fue un extremadamente poderoso huracán de Cabo Verde que causó una destrucción generalizada en su camino a principios de septiembre de 2017. Irma fue el primer huracán de categoría 5 en golpear las Islas de Barlovento, seguido por María dos semanas después.
##SENTENCE##: El huracán Irma fue un extremadamente poderoso huracán de Cabo Verde que causó una destrucción generalizada en su camino a principios de septiembre de 2017.
##DECOMPOSITION##:
```json
[
    {"claim": "Hurricane Irma was a Cape Verde hurricane"},
    {"claim": "Hurricane Irma was extremely powerful"},
    {"claim": "Hurricane Irma caused widespread destruction"},
    {"claim": "Hurricane Irma occurred in early September 2017"}
]
```

##PARAGRAPH## [paragraph]
##SENTENCE## [sentence]
##DECOMPOSITION##:
\end{lstlisting}
\caption{\textbf{Cross-lingual decomposition.} Decomposes a non-English sentence directly into English claims, with no separate translation step.}
\label{prompt:cross-decomp}
\end{figure*}

\begin{figure*}[t]
\begin{lstlisting}[style=prompt]
Instructions:
- You are given a sentence that may be in any language
- Translate it into English, preserving meaning exactly
- Output only the translated sentence, nothing else

##SENTENCE##: Le Brussels Basketball est un club belge de basket-ball fondé en 1957, basé à Bruxelles.
##TRANSLATION##: Brussels Basketball is a Belgian basketball club founded in 1957, based in Brussels.

##SENTENCE##: El huracán Irma fue un extremadamente poderoso huracán de Cabo Verde que causó una destrucción generalizada en su camino a principios de septiembre de 2017.
##TRANSLATION##: Hurricane Irma was an extremely powerful Cape Verde hurricane that caused widespread destruction along its path in early September 2017.

##SENTENCE##: [sentence]
##TRANSLATION##:
\end{lstlisting}
\caption{\textbf{Sentence-translation prompt.} Translates a
source-language sentence into English prior to decomposition.}
\label{prompt:translate-sentence}
\end{figure*}

\begin{figure*}[t]
\begin{lstlisting}[style=prompt]
You are a fact-checker. You are given a source sentence (which may be in any language) and one English claim derived from that sentence.

Determine whether the claim is supported by the source sentence using your multilingual understanding.

Verdict options:
- SUPPORTED: The source sentence explicitly supports the claim.
- CONTRADICTED: The claim states something that directly contradicts the source sentence (wrong fact, wrong name, wrong number, etc.).
- HALLUCINATED: The claim contains information that is not present in the source sentence and cannot be inferred from it.
- CANNOT_DETERMINE: The source sentence does not contain enough information to judge the claim.

Respond with a JSON object only:
{"verdict": "<SUPPORTED|CONTRADICTED|HALLUCINATED|CANNOT_DETERMINE>", "explanation": "<one sentence>"}

##SOURCE SENTENCE##: Brussels Basketball is a professional basketball club based in Brussels, Belgium.
##CLAIM (English)##: Brussels Basketball is a basketball club.
##VERDICT##: {"verdict": "SUPPORTED", "explanation": "The source sentence explicitly states it is a basketball club."}

##SOURCE SENTENCE##: Brussels Basketball is a professional basketball club based in Brussels, Belgium.
##CLAIM (English)##: Brussels Basketball was founded in 1899.
##VERDICT##: {"verdict": "CONTRADICTED", "explanation": "The source sentence does not mention a founding year of 1899; this contradicts information from other knowledge."}

##SOURCE SENTENCE##: Brussels Basketball is a professional basketball club based in Brussels, Belgium.
##CLAIM (English)##: Brussels Basketball has won three national championships.
##VERDICT##: {"verdict": "HALLUCINATED", "explanation": "The source sentence contains no information about championships won."}

##SOURCE SENTENCE##: Le Brussels Basketball est un club belge de basket-ball fondé en 1957, basé à Bruxelles.
##CLAIM (English)##: Brussels Basketball is a Belgian basketball club.
##VERDICT##: {"verdict": "SUPPORTED", "explanation": "The French source sentence states it is a Belgian basketball club (club belge de basket-ball)."}

##SOURCE SENTENCE##: Le Brussels Basketball est un club belge de basket-ball fondé en 1957, basé à Bruxelles.
##CLAIM (English)##: Brussels Basketball was founded in 1962.
##VERDICT##: {"verdict": "CONTRADICTED", "explanation": "The source sentence states the club was founded in 1957 (fondé en 1957), not 1962."}

##SOURCE SENTENCE##: [sentence]
##CLAIM (English)##: [claim]
##VERDICT##:
\end{lstlisting}
\caption{\textbf{LLM-judge prompt.} Fact-checks an English claim against a (possibly non-English) source sentence, returning one of four verdicts.}
\label{prompt:claim-support}
\end{figure*}

\begin{figure*}[t]
\begin{lstlisting}[style=prompt]
You audit whether a GOLD ANSWER is a FAIR grading key for a quiz question — i.e. would a knowledgeable, correct responder be EXPECTED to produce this specific answer, or could a correct response reasonably differ? You are given the SOURCE CLAIM the QA pair was written from, for reference.

GOOD: the gold is the clear, essentially-unique expected answer. A correct responder would produce it (allowing trivial phrasing/format differences). Grading against it is fair.

QUESTIONABLE: grading against the gold could wrongly penalize a correct answer — because the question admits SEVERAL equally valid answers and the gold is just one of them, the gold is vague/under-specified, or it is an oddly-specific / source-dependent phrasing where a correct responder might reasonably give a different valid answer.

BAD: the gold is wrong, nonsensical, contradicts the source claim, or does not actually answer the question.

Examples:
Claim: The company was founded in 1923. | Q: In what year was the company founded? | Gold: 1923 -> GOOD
Claim: The capital is Paris. | Q: What is the capital? | Gold: Paris -> GOOD
Claim: The capital is Paris. | Q: What is the capital? | Gold: a large European city -> BAD
Claim: The bridge is 110 km long. | Q: How long is the bridge? | Gold: 145 km -> BAD
Claim: Amulets have parallels to votive offerings, talismans, and charms. | Q: What do amulets have parallels to? | Gold: votive offerings -> QUESTIONABLE
Claim: He was known for his kindness, patience, and humor. | Q: What was he known for? | Gold: patience -> QUESTIONABLE
Claim: The team won titles in 1998, 2001, and 2004. | Q: When did the team win a title? | Gold: 2001 -> QUESTIONABLE

Reply with ONLY one label: GOOD, QUESTIONABLE, or BAD.
\end{lstlisting}
\caption{\textbf{Gold-answer fairness prompt.} Judges whether a gold answer is a
fair grading key for a quiz item (GOOD / QUESTIONABLE / BAD), given the source
claim it was written from.}
\label{prompt:gold-audit}
\end{figure*}

\begin{figure*}[t]
\begin{lstlisting}[style=prompt]
You judge whether a quiz question can be answered STANDALONE — without access to the one specific hidden source document it was written from.

CONTEXT_DEPENDENT: cannot be resolved without that source. It refers to an entity by a pronoun or a bare definite phrase with no unique identifier ("the model", "the company", "the ship", "the game"), names a person only by a common surname, or presumes one specific unnamed work/game/organization/place.

SELF_CONTAINED: the subject is uniquely identifiable from the question alone — a proper noun or precise description that a knowledgeable person or a web search could pin down to exactly one thing.

BORDERLINE: it names a specific-sounding entity, but resolving it still needs outside knowledge of a country/domain (e.g. a generically-named council or body), OR the subject is clear but the requested answer is under-determined (many valid answers).

Examples:
Q: Who is the captain of the ship? -> CONTEXT_DEPENDENT
Q: What did she regret in the final scene? -> CONTEXT_DEPENDENT
Q: What year did Smith join the company? -> CONTEXT_DEPENDENT
Q: What is an example of a special ability in the game? -> CONTEXT_DEPENDENT
Q: When did Apollo 11 land on the Moon? -> SELF_CONTAINED
Q: How tall is the Eiffel Tower? -> SELF_CONTAINED
Q: Who wrote the novel Pride and Prejudice? -> SELF_CONTAINED
Q: Which industries does the National Fisheries Council oversee? -> BORDERLINE
Q: What is the primary export of the region? -> BORDERLINE

Reply with ONLY one label: SELF_CONTAINED, BORDERLINE, or CONTEXT_DEPENDENT.
\end{lstlisting}
\caption{\textbf{Standalone-answerability prompt.} Judges whether a quiz question can be answered without its one hidden source document (SELF\_CONTAINED / BORDERLINE / CONTEXT\_DEPENDENT).}
\label{prompt:question-audit}
\end{figure*}

\begin{figure*}[t]
\begin{lstlisting}[style=prompt]
You are reviewing proposed additions to the English Wikipedia article "[article_title]". Each numbered claim below was extracted from a non-English edition of the same article. For each claim, determine its relationship to the ARTICLE TEXT:

- SUPPORTED — the article already states this information, either directly or through a more specific statement that entails it. The claim adds nothing new.
- ABSENT — the article does not state this information, and nothing in the article conflicts with it. The claim is a candidate addition.
- CONTRADICTED — the article states something incompatible with the claim (a different date, number, name, place, quantity, or relationship).

Rules:
- Judge ONLY against the article text below. Do not use outside knowledge about the topic.
- A claim that is partially in the article: if the new part conflicts with the article, use CONTRADICTED; if the new part is simply not mentioned, use ABSENT.
- For SUPPORTED and CONTRADICTED, copy the single most relevant sentence or span from the article verbatim into "evidence" (at most 40 words). For ABSENT, set "evidence" to null.

ARTICLE TEXT:
[article_text]

CLAIMS:
[claims]

Respond with ONLY a JSON array, one object per claim, in order:
[{"idx": 1, "verdict": "SUPPORTED", "evidence": "..."}, {"idx": 2, "verdict": "ABSENT", "evidence": null}, ...]
\end{lstlisting}
\caption{\textbf{Claim-review.} Classifies each candidate claim (extracted from a non-English edition) against the full English article text as SUPPORTED, ABSENT, or CONTRADICTED.}
\label{prompt:claim-review}
\end{figure*}

\begin{figure*}[t]
\begin{lstlisting}[style=prompt]
You are editing the English Wikipedia article "[title]". Below is the article's current section outline, then a list of new facts (claims) to add to the article. For EACH claim, choose the single existing section it best belongs in. If no existing section is a good fit, assign it 0 (a new section will be created for it).

Prefer an existing section — only use 0 when the claim's topic is genuinely not covered by any section.

SECTION OUTLINE:
[outline]

CLAIMS:
[claims]

Respond with ONLY a JSON array, one object per claim, in order:
[{"idx": 1, "section": 3}, {"idx": 2, "section": 0}, ...]
where "section" is the [n] index of the chosen section, or 0 for "needs a new section".
\end{lstlisting}
\caption{\textbf{Section-routing.} Assigns each new claim to the existing section it best fits, or to 0 when no existing section is suitable and a
new one is needed.}
\label{prompt:route-claims}
\end{figure*}

\begin{figure*}[t]
\begin{lstlisting}[style=prompt]
You are editing the English Wikipedia article "[title]". The claims below did not fit any existing section (outline shown for context). Propose one or more NEW sections to hold them, grouping related claims together. Keep the number of new sections small; only propose a section when several claims share a topic, or a single claim is clearly its own topic.

EXISTING OUTLINE:
[outline]

ORPHAN CLAIMS:
[claims]

Respond with ONLY a JSON array of proposed sections:
[{"heading": "Reception", "after_section": 4, "claim_idxs": [1, 3, 5]}, ...]
- "heading": the new section title (Wikipedia style, plain).
- "after_section": the [n] index of the existing section this new section should follow (use the highest sensible index; 0 to place right after the lead).
- "claim_idxs": the claim numbers (from the list above) that belong in this section.
Every claim must appear in exactly one proposed section.
\end{lstlisting}
\caption{\textbf{New-section planning prompt.} Groups the orphan claims that fit no existing section into a small set of proposed new sections, each with a heading and insertion point.}
\label{prompt:route-new-sections}
\end{figure*}

\begin{figure*}[t]
\begin{lstlisting}[style=prompt]
You are editing the "[heading]" section of the English Wikipedia article "[title]". Integrate the NEW FACTS into the section as FLUENT, natural encyclopedic prose.

Rules — WRITE WELL, DON'T LIST:
- When several new facts share a subject, MERGE them into a single flowing sentence using coordination and lists — do NOT write one short sentence per fact.
  Example — these four facts:
    - The 1876 Constitution maintained Spain as a constitutional monarchy.
    - The 1876 Constitution granted the king the power to appoint members of the Senate.
    - The 1876 Constitution granted the king the power to repeal laws.
    - The 1876 Constitution granted the king the title of Commander-in-Chief of the Army.
  should become ONE sentence:
    "The 1876 Constitution maintained Spain as a constitutional monarchy, granting the king the power to appoint members of the Senate and repeal laws, as well as the title of Commander-in-Chief of the Army."
- Keep existing sentences as-is unless a new fact must be woven in; place new prose next to the related existing content. Do not drop existing information; do not invent anything beyond the facts.
- CITATIONS: some facts end with a tag like [cite: 3]. Append the marker(s) at the END of the sentence that uses the fact ("...Army.[3]"); combine when merging cited facts ("...[3][5]"). Use only numbers given; never invent; never write the "[cite: ...]" text.
- Objective, encyclopedic tone. Output ONE SENTENCE PER LINE (a merged multi-fact sentence is ONE line). Blank line between paragraphs. No heading, no commentary.

CURRENT SECTION (one sentence per line):
[current]

NEW FACTS TO ADD:
[claims]

Respond with ONLY the rewritten section text, one sentence per line.
\end{lstlisting}
\caption{\textbf{Fluent section-rewrite prompt.} Integrates new facts into an existing section as natural prose, merging facts that share a subject into single flowing sentences rather than one sentence per fact.}
\label{prompt:rewrite}
\end{figure*}

\begin{figure*}[t]
\begin{lstlisting}[style=prompt]
You are writing the "[heading]" section of the English Wikipedia article "[title]" using ONLY the facts below. Write FLUENT, natural encyclopedic prose — not a list of one-fact sentences.

Rules — WRITE WELL, DON'T LIST:
- MERGE facts that share a subject into single flowing sentences using coordination and lists.
  Example — these four facts:
    - The 1876 Constitution maintained Spain as a constitutional monarchy.
    - The 1876 Constitution granted the king the power to appoint members of the Senate.
    - The 1876 Constitution granted the king the power to repeal laws.
    - The 1876 Constitution granted the king the title of Commander-in-Chief of the Army.
  should become ONE sentence:
    "The 1876 Constitution maintained Spain as a constitutional monarchy, granting the king the power to appoint members of the Senate and repeal laws, as well as the title of Commander-in-Chief of the Army."
- Use only the given facts; do not invent detail or add outside knowledge. Group related facts into coherent sentences and paragraphs.
- CITATIONS: some facts end with a tag like [cite: 3]. Append the marker(s) at the END of the sentence ("...Army.[3]"); combine when merging ("...[3][5]"). Use only numbers given; never invent; never write the "[cite: ...]" text.
- Objective, encyclopedic tone. Output ONE SENTENCE PER LINE (a merged multi-fact sentence is ONE line). Blank line between paragraphs. No heading, no commentary.

FACTS:
[claims]

Respond with ONLY the section text.
\end{lstlisting}
\caption{\textbf{Fluent new-section prompt}. Drafts a new section from only
its facts as natural prose, merging related facts into flowing sentences.}
\label{prompt:rewrite-new-section}
\end{figure*}

\begin{figure*}[t]
\begin{lstlisting}[style=prompt]
A user requested the following report:

REPORT REQUEST:
[request]

Below are candidate facts extracted from source documents. For EACH fact, decide whether it is ON-TOPIC and relevant to include in THIS report — i.e. it directly addresses the report request or provides context a reader of this report would need. Mark a fact NOT relevant if it is off-topic, about an unrelated event/entity, generic boilerplate, or would not belong in this specific report even if true.

Judge relevance to the request only; do not judge whether the fact is interesting or novel.

CANDIDATE FACTS:
[claims]

Respond with ONLY a JSON array, one object per fact, in order:
[{"idx": 1, "relevant": true}, {"idx": 2, "relevant": false}, ...]
\end{lstlisting}
\caption{\textbf{Query-relevance prompt.} Judges whether each candidate fact is
on-topic for the requested report (relevant true/false), independent of whether
the fact is interesting or novel.}
\label{fig:prompt-query-relevance}
\end{figure*}